\documentclass{bmvc2k}

\usepackage{multirow}
\usepackage{cleveref}
\usepackage{makecell}

\usepackage[labelformat=simple]{subcaption}

\crefname{figure}{Fig.}{Fig.}
\crefname{equation}{Eq.}{Eq.}
\crefname{table}{Table}{Table}
\crefname{section}{Section}{Section}

\title{Capturing Temporal Information in a Single Frame: Channel Sampling Strategies for Action Recognition}

\addauthor{Kiyoon Kim}{kiyoon.kim@ed.ac.uk}{}
\addauthor{Shreyank N Gowda}{s.narayana-gowda@sms.ed.ac.uk}{}
\addauthor{Oisin Mac Aodha}{oisin.macaodha@ed.ac.uk}{}
\addauthor{Laura Sevilla-Lara}{lsevilla@ed.ac.uk}{}

\addinstitution{
 School of Informatics \\
 University of Edinburgh \\
 Edinburgh, UK
}

\runninghead{Kim et al.}{Channel Sampling Strategies for Action Recognition}

\def\eg{\emph{e.g}\bmvaOneDot}

\def\ie{\emph{i.e}\bmvaOneDot}

\def\etc{\emph{etc}\bmvaOneDot}

\usepackage{fp}
\newcommand{\roundnum}[1]{\FPeval{\tempnum}{round(#1,1)}\tempnum}

\newcommand{\paragraphbf}[1]{\noindent \textbf{#1}\hspace{1em}}

\def\@fnsymbol#1{\ensuremath{\ifcase#1\or *\or \dagger\or \ddagger\or
   \mathsection\or \mathparagraph\or \|\or **\or \dagger\dagger
   \or \ddagger\ddagger \else\@ctrerr\fi}}
\newcommand{\nsymbol}[1]{\@fnsymbol{#1}}
\newcommand{\ssymbol}[1]{$^{\@fnsymbol{#1}}$}

\begin{document}

\maketitle

\begin{abstract}
We address the problem of capturing temporal information for video classification in 2D networks, without increasing their computational cost. 
Existing approaches focus on modifying the architecture of 2D networks (\eg by including filters in the temporal dimension to turn them into 3D networks, or using optical flow, \etc), which increases computation cost. Instead, we propose a novel sampling strategy, where we re-order the channels of the input video, to capture short-term frame-to-frame changes. We observe that without bells and whistles, the proposed sampling strategy improves performance on multiple architectures (\eg TSN, TRN, TSM, and MVFNet) and datasets (CATER,  Something-Something-V1 and V2), up to 24\% over the baseline of using the standard video input.
In addition, our sampling strategies do not require training from scratch and do not increase the computational cost of training and testing. Given the generality of the results and the flexibility of the approach, we hope this can be widely useful to the video understanding community. Code is available on our website: \href{https://github.com/kiyoon/channel\_sampling}{https://github.com/kiyoon/channel\_sampling}.
\end{abstract}

\section{Introduction}

Understanding temporal information is crucial in order to understand video. While 2D convolutional filters are usually the standard approach to capture spatial information, there is a wider variety of methods for capturing temporal information. These include, for example, using the transformer architecture~\cite{timesformer} 3D networks~\cite{c3d}, computing optical flow and feeding it to a 2D convolutional network~\cite{twostream}, using relational networks~\cite{trn}, or using Recurrent Neural Networks (RNNs)~\cite{faster}, among others. All of these methods require changes in the underlying network architectures, additional computational cost compared to simple 2D networks, as well as the time-consuming process of pre-training from scratch.

In this paper, we propose two simple and novel channel sampling strategies that improve the ability of a given 2D network to capture temporal information without changing the architecture. In particular, we re-order the channels of the input video in two ways: in the first, we re-order the channels of the video, so that %
each frame is composed of three channels: one that belongs to the frame before, one that belongs to the current frame, and one that belongs the frame after. These three channels are concatenated in the channel dimension as if it were a single frame. This procedure incorporates temporal information from neighboring frames, while keeping the dimensions of the input frame. %
\begin{figure}[t]
    \centering
    \includegraphics[width=0.6\columnwidth]{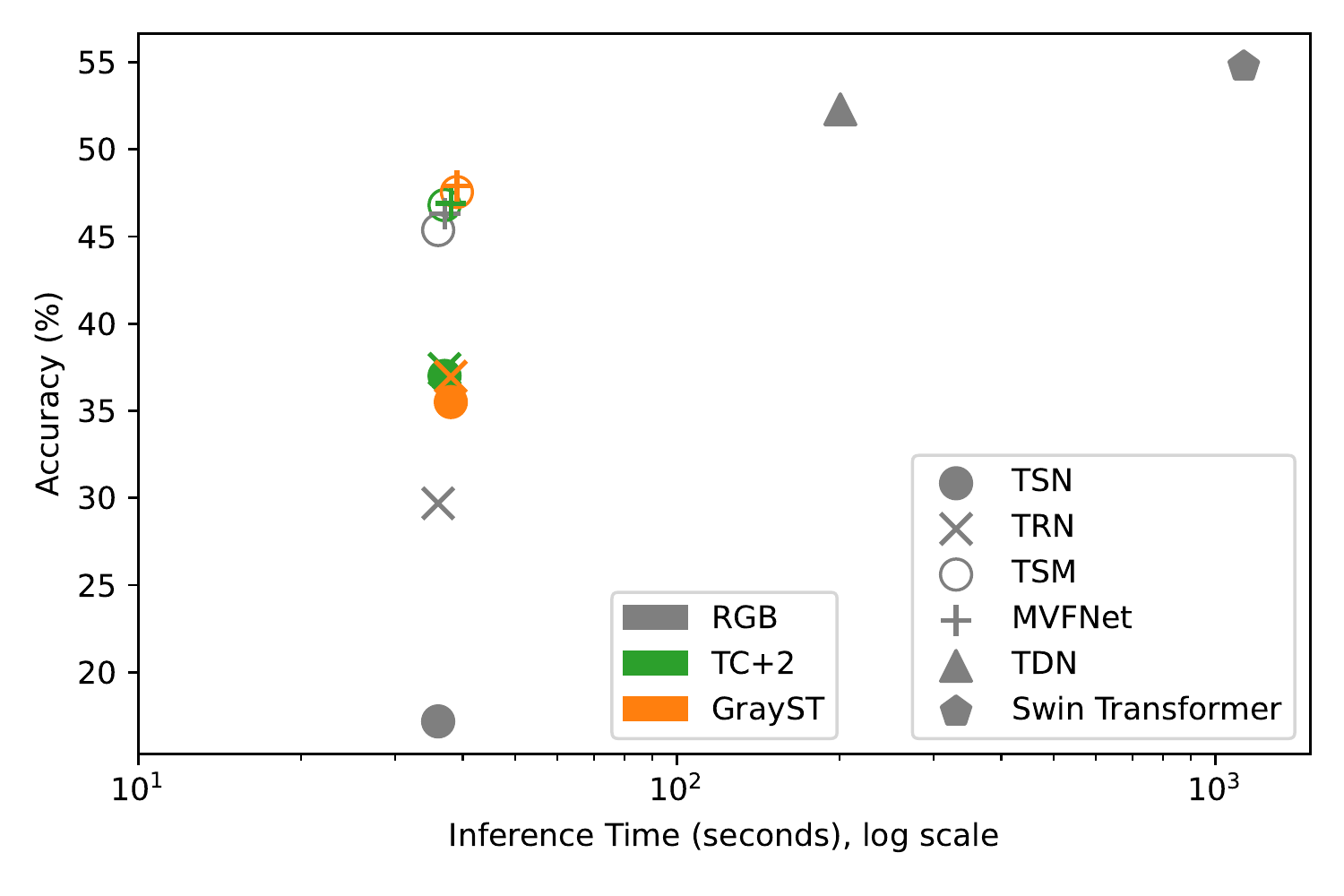}
    \vspace{-10pt}
    \caption{
    We show that it is possible to significantly increase the performance of lightweight action recognition networks on the challenging Something-Something-V1 dataset by simply adapting how the individual image channels are sampled. 
    Our approaches, \emph{TC+2} (in green) and \emph{GrayST} (in orange), improve accuracy across several networks without increasing inference time. 
    This enables us to narrow the gap between these efficient baselines and much more computationally demanding methods such as TDN~\cite{tdn} and Video Swin Transformers~\cite{swin}. 
    Note that we use a log scaling on the horizontal axis.
    }
    \label{fig:accuracy_vs_inference_time}
    \vspace{-10pt}
\end{figure} 
In the second strategy, we compute a grayscale version of three neighboring frames at a time, and again concatenate them in the channel dimension, as if they were a single RGB image. 

We test these extremely simple re-ordering strategies on five widely used 2D networks (TSN~\cite{tsn}, TRN/MTRN~\cite{trn}, TSM~\cite{tsm}, and MVFNet~\cite{wu2021mvfnet}). 
We observe that without any additional engineering these re-ordering strategies improve results up to 24\% compared to the standard RGB channel ordering, across multiple challenging video datasets and different networks. 
We also observe that the improvement is particularly large in the datasets where temporal information is more important, in particular on CATER~\cite{cater} and Something-Something~\cite{something}. Figure~\ref{fig:accuracy_vs_inference_time} illustrates the performance improvement which does not add additional significant computational overhead.

Our main contributions are:
\begin{itemize}
    \vspace{-7pt}
    \item We improve the performance of simple 2D CNN-based action recognition models while incurring no additional computational complexity and with no modification to the underlying models. Our solutions can be used with most existing network architectures.
    \vspace{-7pt}
    \item Through extensive experimental evaluation on several common models and datasets,  we show the efficiency of our proposed sampling methods and report superior performance compared to standard image sampling.
\end{itemize}

\section{Related Work}
\paragraphbf{Representations of Input Video.}
The most common representation of video frames for deep learning is 8-bit raw RGB, where each pixel has three color channels -- red, green, and blue, each represented by 8 bits. This representation does not encode any temporal information in a frame. Another approach is to compute differential images by simply subtracting each pair of consecutive frames \cite{tsn}.
This gives the network a more explicit representation of temporal high-frequency changes. Unfortunately, this is not robust to spatial shifting, so as the camera pans, all pixel values change, which defeats the purpose of identifying the moving objects by subtracting the frames.

Optical flow is another common representation in action recognition \cite{twostream,tsn,i3d}.
Given two consecutive video frames, the optical flow is the direction and magnitude of the motion at each pixel. 
Using optical flow as a separate stream~\cite{twostream} generally improves performance, although it is often computationally expensive. Recent work has used the motion vectors from compressed video (\eg H.264) to avoid the computational cost of optical flow~\cite{wu2018compressed,zhang2016real}. Given any of these representations, one of the standard approaches is to simply feed them as input to a 2D convolutional network, similar to those used in the image domain, where frames are processed independently, and the predictions are aggregated in the end. This family of approaches is called 2D, or frame-based networks. Despite their simplicity, they often work well on many datasets, and are fast to train and test. 

Dynamic images \cite{bilen2017action} are a compact video representation that encode spatial and the entire motion information in a single image, so any image classification network can perform action recognition using the generated images. 
In contrast, our methods encode short-term temporal information per frame, resulting in space-time sequential representation (\ie they do not compress an entire video into a single frame) and can thus use off-the-shelf 2D action recognition networks to better capture temporal information compared to only using a single image classification model as in \cite{bilen2017action}.
\\

\paragraphbf{3D Architectures.} The main issue with 2D networks is that they are limited in their temporal receptive field size. 
3D networks \cite{c3d} on the other hand, benefit from learning spatio-temporal representations explicitly as they apply convolutions in both space and time. One of the most popular approaches is the I3D model \cite{i3d} that inflates 2D models to 3D. %
As a result, they are able to make use of well established 2D architectures and pre-trained weights. 
Extensions to I3D include adding a non-local (NL) module~\cite{nonlocal} to help capture pixel-level correlations, resulting in  improved performance. 
The SlowFast network \cite{slowfast} is also a widely used 3D network, and it uses two pathways, a slow and a fast one, to capture motion and fine temporal information. 
While 3D networks are helpful, they are much more expensive than their 2D counterparts. 
In this work we aim to introduce temporal information into 2D backbone-based models without requiring the computation of optical flow or without expensive 3D convolutions. 
\\

\paragraphbf{2D Architectures for Temporal Modeling.}
While 3D action recognition networks can perform better than their 2D counterparts, they have larger computational requirements for training and testing. 
This in turn, makes them challenging to deploy in resource-constrained settings, \eg mobile and online scenarios. 
As a result, 2D networks for action recognition tasks is still common practice for certain application domains. 

Temporal Segment Networks (TSN)~\cite{tsn} is an early and widely used 2D method. 
TSN simply extracts predictions for each frame sampled from the input video using a 2D backbone network and then averages these predictions for the final output. 
Frames are sparsely sampled from longer videos so that the model can reason over long time spans. 
With an added optical flow stream, TSN achieves competitive performance on action datasets, even defeating some 3D networks at the time of release. 
This high performance is more obvious in datasets where there is a strong correlation between actions and static objects and scenes~\cite{kinetics}. 
In contrast, TSN performs worse on datasets that require explicit temporal reasoning~\cite{something,onlytime}.

Temporal Relation Networks (TRN)~\cite{trn} aim to capture relational information across frames. For this, it also uses a 2D backbone to extract features from sparsely sampled frames, but then aggregates the frame-level features using a relational module \cite{relationnet}. 
The authors proposed a multi-scale relational module that learns 2-frame, 3-frame, and up to $T$-frame relationships (TRN-multiscale, or MTRN in short). This slightly improves performance over the single-scale TRN. 
This is an improvement over TSN, which is agnostic to the temporal ordering of the input frames. 
As a result, TRN improves performance on temporal datasets, \eg \cite{something}.
However, their model is not very memory-efficient, as it requires $T-1$ relational modules (from 2 to $T$ frame) using fully-connected layers, making it challenging to process many frames.

The Temporal Shift Module (TSM)~\cite{tsm} follows the same strategy as the TSN but modifies the ResNet~\cite{resnet} backbone so that the network can partially access information from one past and one future frame (\ie temporal shift).
In order to maintain the spatial feature learning capability, the temporal shift happens inside a residual branch, so the backbone choice is limited to those with residual connections.
TSM achieved improved performance on the Something-Something-V2 dataset \cite{something}, showing that 2D models are not obsolete and are still competent in action recognition.

Multi-View Fusion Network (MVFNet) \cite{wu2021mvfnet} adds channel-wise convolutions to model height-width, height-time, and width-time views instead of treating height-width-time video frames as a space-time signal. This approach performs better than TSM while still remaining computation efficient.

Finally, Temporal Difference Networks (TDN)~\cite{tdn} explicitly compute motion information by sampling five times as many input frames (\eg 8-frame TDN uses 40 frames) and computes RGB differences for capturing short-term information and uses multi-scale attention for capturing long-term temporal information.  TDN presented a new state-of-the-art on the Something-Something datasets, but at the cost of using significantly more input frames.

In this work, we take advantage of the efficiency and simplicity of 2D video models by proposing image channel sampling strategies that improve a model's ability to capture temporal information. 
This increases accuracy, all without introducing any additional computation during training or testing.

\section{Method} 

\subsection{Overview}

Given an input video $v$, the task of action recognition is to predict the action class label $y$ of the video. %
Typically, the input video $v$ can be sampled sparsely, which means that each video is evenly divided into $T$ segments. 
Then, in the standard 2D paradigm, a frame $\mathbf{x}_t$ is chosen at random for each segment at training time. %
We call the set of sampled frames $\mathbf{X}$, \ie $\mathbf{X} = \left\{\mathbf{x}_1, \mathbf{x}_2, \dots, \mathbf{x}_T\right\}$.

We let $f(\mathbf{x}_i)$ denote the 2D backbone model that takes a single image as input, and outputs a feature map. We compute $f(\mathbf{X}) = \left \{f(\mathbf{x}_1), f(\mathbf{x}_2), \dots, f(\mathbf{x}_T)\right \}$. Finally, $g(\cdot)$ is a temporal aggregation module that takes the $T$ feature maps and returns the output category prediction $\hat{y}$. 
The aggregation function $g(\cdot)$ can simply be an average of the individual predictions as in TSN~\cite{tsn,tsm},

\begin{equation}
g(f(\mathbf{X})) = \frac{\sum_{t=1}^T{h_\theta(f(\mathbf{x}_t))}}{T},
\end{equation}
where $h_\theta$ is a per-frame classifier.  
In other cases, like TRN~\cite{trn}, the aggregation function reasons about the relationship among multiple frames by concatenating the features and applying a multi-layer perceptron (MLP) $h_\phi$,

\begin{equation}
    g(f(\mathbf{X})) = h_\phi(\text{concat}(f(\mathbf{X}))).
\label{equ:TRN}
\end{equation}
The multi-scale version of TRN (MTRN) works similarly but has several aggregation modules, each accounting for reasoning between 2-frame, 3-frame, \dots, $T$-frame relationships. %

Across variants of the 2D paradigm, the frames of a video are typically sampled sequentially with all color channels intact. 
Instead, we propose two alternative sampling strategies, \emph{Time-Color Reordering} and \emph{Grayscale Short-Term Stacking} that enables 2D backbone-based models to exploit temporal information by means of reordering the input channels.
Despite their conceptual simplicity, these two sampling strategies significantly improve action recognition performance without requiring any structural changes to the backbone model. 

\begin{figure*}
\centering

\begin{subfigure}[t]{0.8\textwidth}
    \centering
    \includegraphics[width=\textwidth]{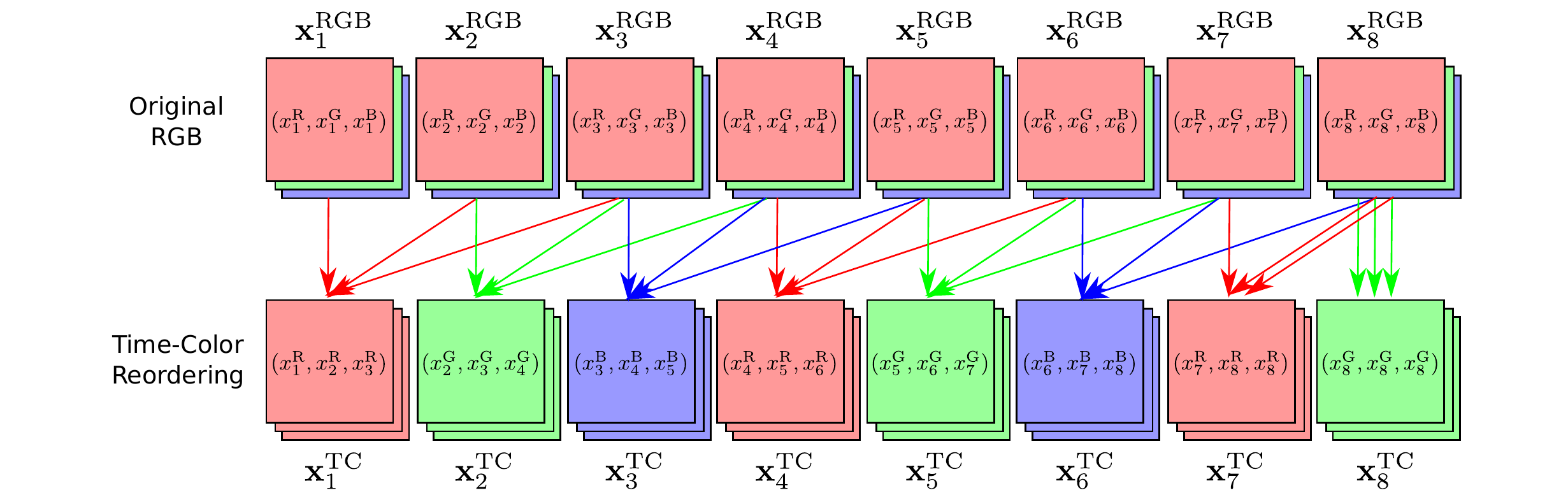}
    \caption{8-frame TC Reordering}
    \label{fig:8frame_TC}
\end{subfigure}

\begin{subfigure}[t]{0.8\textwidth}
    \centering
    \includegraphics[width=\textwidth]{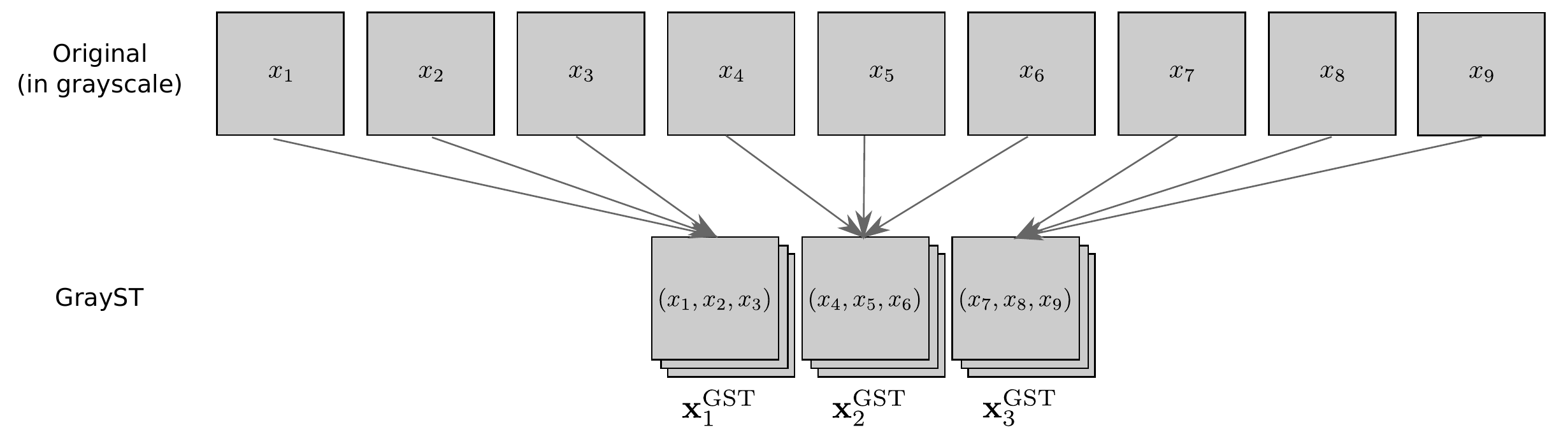}
    \caption{3-frame GrayST. The input video frames are converted to grayscale and the output is simply the concatenation of these frames into groups of three channels} 
    \label{fig:3frame_GrayST}
\end{subfigure}

\vspace{7pt}
\caption{Visualization of our TC Reordering \subref{fig:8frame_TC} and GrayST \subref{fig:3frame_GrayST} methods.
In each case, the top row represents the original input frames, \ie eight RGB frames in \subref{fig:8frame_TC}, and nine grayscale frames in \subref{fig:3frame_GrayST}. 
Note also that, GrayST gets to use three times as many input frames, \eg to generate 8 output frames, one needs to sample 24 inputs. 
}
\label{fig:TC}
\vspace{-10pt}
\end{figure*}

\begin{figure*}
\centering
\includegraphics[width=0.8\textwidth]{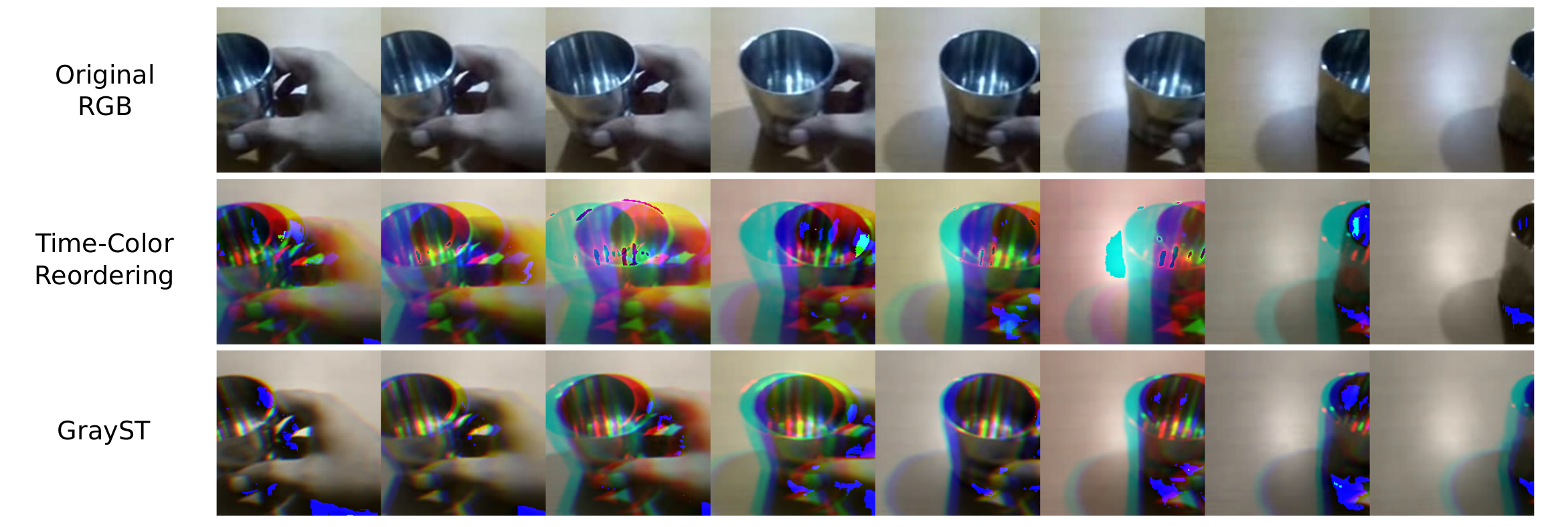}
\vspace{-5pt}
\caption{Visualization of the different sampling strategies. All frames are interpreted as if they were RGB. The video depicted is an instance of ``Pulling something from left to right'' from the Something-Something-V2 dataset. Our two sampling strategies make the motion and its direction clearly visible by utilizing three color channels.} 
\label{fig:sampling_visualisation}
\vspace{-10pt}
\end{figure*}

\subsection{Time-Color Reordering}
\label{sec:tc_overview}
We propose a simple yet powerful sampling technique for video frames called \emph{Time-Color (TC) Reordering}. We allow the 2D backbone model $f(\mathbf{x}_i)$ to see
temporal information by re-sampling the three color channels of the model. 
We do this by taking one color channel from the input clip (\ie red) of 3 consecutive frames, and concatenating in the channel dimension to form an input ``image". Then we repeat the process with the next color channel (\ie green), and so on.
Note that this is merely a different representation of the input video by means of changing the channel order, and does not require any modification to the backbone model or include additional information or processing. More specifically, the process is as follows. 
Let $\mathbf{X}^{\text{TC}} = \left\{\mathbf{x}_1^{\text{TC}}, \mathbf{x}_2^{\text{TC}}, \dots, \mathbf{x}_T^{\text{TC}}\right\}$ denote a video clip sampled following the proposed TC Reordering sampling strategy, 
where
\begin{equation}
  \mathbf{x}_i^\text{TC}=\begin{cases}
    (x_i^\text{R}, x_{i+1}^\text{R}, x_{i+2}^\text{R}), & \text{if $i$ mod 3 = 1}.\\
    (x_i^\text{G}, x_{i+1}^\text{G}, x_{i+2}^\text{G}), & \text{if $i$ mod 3 = 2}.\\
    (x_i^\text{B}, x_{i+1}^\text{B}, x_{i+2}^\text{B}), & \text{otherwise}.\\
  \end{cases}
\end{equation}
Here, $x_i^\text{R}$ is the red channel from the $i$-th frame, $x_i^\text{G}$ for green, and $x_i^\text{B}$ for blue. Since the last two TC frames try to access future frames (\ie $x_{T+1}$ and $x_{T+2}$) that are not available, we just use $x_T$ with the corresponding color channel for this case. This simple process yields frames that contain information about the neighboring frames, and it is the core reason for the large advantage we observe in temporal tasks.  \cref{fig:8frame_TC} outlines the procedure in detail. 

We alternate color channels to expose the network to more varied data. That is, different channels depict a particular object with different brightness and contrast, while keeping the original shape of the object. We observe that this color alternation produces better results than a single one, and speculate that it may work as a form of data augmentation.

We also propose an alternative method \emph{TC+2} where we sample just two more frames to avoid the duplication of the last frames. The cost of sampling two more frames is almost negligible in many practical scenarios even with long-input networks, and we observe significant improvement of performance as the way the input data is formed is consistent throughout the frames.

\subsection{Grayscale Short-Term Stacking}
Here we introduce another sampling technique that we call \emph{Grayscale Short-Term Stacking}  (\emph{GrayST}). 
This approach is designed to use more source frames with the same compact 2D networks by using grayscale images instead of RGB.
The motivation is that in semantic understanding tasks, color information can be redundant, and we can use that capacity to include temporal information instead, by inputting more frames. 
Previous work~\cite{xie2018pre} showed that there is only a 0.5\% drop in accuracy on ImageNet~\cite{imagenet}, when training on grayscale images compared to RGB images. 
Thus, we replace the three color channels that are normally fed into a 2D backbone network with three grayscale frames, from three different sequential time steps. 
In effect, this allows the backbone model to see short temporal information at the expense of forgoing the ability to reason about color. 
This sampling strategy is visualized in \cref{fig:3frame_GrayST}. 

When the input sequence length to a network is intended to be $T$ frames, we simply sample $3\times{}T$ frames in grayscale, and stack the three consecutive frames into one image, containing three grayscale channels. 
In offline scenarios, we can utilize higher temporal resolution per video clip without introducing any latency for training or testing. Some online scenarios may have a limited number of frames, so we also experiment with matching the number of input RGB frames, \ie 8-frame GrayST (that sees 24 frames) vs 24-frame RGB. 

We let $\mathbf{X}^\text{GST} = \left\{\mathbf{x}_1^\text{GST}, \mathbf{x}_2^\text{GST}, \dots, \mathbf{x}_T^\text{GST}\right\}$ denote a GrayST video clip. 
We first sample $3 \times T$ grayscale frames following the same sparse sampling strategy as before,
\begin{equation}
\mathbf{X}^g = \left\{x^g_1, x^g_2, \dots, x^g_{3T}\right\}, 
\end{equation}
where $x^g_i$ is a grayscale image. 
Then, a GrayST frame is made of three neighboring temporal frames in $\mathbf{X}^g$ which is defined as 
\begin{equation}
    \mathbf{x}_i^\text{GST} = (x^g_{3i-2}, x^g_{3i-1}, x^g_{3i}).
\end{equation}
A comparison between the two sampling techniques is visualized in \cref{fig:sampling_visualisation}.

\section{Evaluation}
In this section we evaluate our two channel sampling strategies on multiple different action recognition datasets using several different network architectures.

\subsection{Experiment Details}

\paragraphbf{Datasets.}
We experiment with challenging datasets that require extensive temporal reasoning. 
The datasets are chosen to span a wide range of situations, \eg short and long range temporal reasoning, static and moving cameras, \etc. 
CATER \cite{cater} is a synthetic action recognition dataset involving  long-term temporal reasoning. CATER has two versions of the videos, with camera motion and without, and we experiment with both of them. CATER also consists of three different tasks: primitive multi-label action recognition (task 1), compositional multi-label action recognition (task 2), and localization of object of interest (task 3). We evaluate task 2, compositional action recognition, as it requires the longest temporal reasoning (from the beginning to the end of the videos consisting of 301 frames). 
This provides the biggest challenge for action recognition methods that only focus on short-term clips.

Specifically, CATER task 2 has 301 classes. Each class represents the ordering of two action pairs: an object performing an action before/during/after another object performing another action, for example, ``cone slides before cylinder rotates''. Furthermore, CATER allows containment of objects, and even recursion of it. That is, the action models have to remember which object is contained by which carrier, during which period, to successfully keep track of all actions happening in the scene.

We also evaluate on datasets that require temporal understanding such as Something-Something-V1 and Something-Something-V2 \cite{something}. The Something-Something datasets consist of videos of pre-defined actions being performed using everyday objects. To focus on the action, and not the objects being used, these datasets removed object and scene bias by grouping the same action using various objects. 
The labels include ``pushing something from right to left'', ``putting something on a surface'', \etc, which ensures the focus is on the action. 
Something-Something-V1 contains 108,499 videos with 174 class labels, while Something-Something-V2 consists of 220,847 videos of the same 174 classes.

\paragraphbf{Networks. } 
We use a ResNet50~\cite{resnet}, pre-trained on ImageNet~\cite{imagenet}, as our 2D backbone model for all experiments. 
Both sampling strategies would likely benefit further from the backbone being pre-trained for their specific channel sampling strategies, but we leave this for future work.  
We mainly present results for five popular 2D models: TSN~\cite{tsn}, TRN/MTRN~\cite{trn}, TSM~\cite{tsm}, and MVFNet~\cite{wu2021mvfnet} with some additional experiments using I3D~\cite{i3d} pre-trained on Kinetics-400~\cite{kinetics} for completeness. 
Note, that we were not able to perform 32-frame MTRN experiments due to GPU VRAM limitations, as it has 31 temporal relational modules consisting of dense fully-connected layers. More details on the training settings can be found in the supplementary material.

\subsection{Results}

\begin{table}[t]

\centering
\resizebox{0.45\linewidth}{!}{
\begin{tabular}{|l|l|c|c|}
\hline
\multirow{2}{*}{Model} & \multirow{2}{*}{Sampling} & \multicolumn{2}{c|}{mAP} \\
\cline{3-4}
 && Static & Camera Motion\\
\hline
\hline
\multirow{3}{*}{TSN} & RGB & \roundnum{49.61} & \roundnum{51.62}\\ 
& TC & \textbf{\roundnum{73.73}} & \roundnum{56.60} \\ 
& TC+2 & \roundnum{73.46} & \roundnum{60.50} \\ 
& GrayST & \roundnum{71.93} & \textbf{\roundnum{61.85}} \\ 
\hline
\multirow{3}{*}{TRN} & RGB & \roundnum{54.86} & \roundnum{54.71}\\ 
& TC & \textbf{\roundnum{72.36}} & \roundnum{52.89} \\ 
& TC+2 & \roundnum{72.32} & \roundnum{52.90} \\ 
& GrayST & \roundnum{69.80} & \textbf{\roundnum{57.59}}\\ 
\hline
\multirow{3}{*}{TSM} & RGB & \roundnum{79.89} & \roundnum{65.75}\\ 
& TC & \roundnum{81.21} & \roundnum{63.25}\\ 
& TC+2 & \roundnum{82.02} & \roundnum{63.98}\\ 
& GrayST & \textbf{\roundnum{82.24}} & \textbf{\roundnum{74.74}}\\ 
\hline
\multirow{3}{*}{MVFNet} & RGB & \roundnum{80.18} & \roundnum{63.48}\\ 
& TC & \roundnum{82.13} & \roundnum{62.65}\\ 
& TC+2 & \roundnum{82.76} & \roundnum{65.45}\\ 
& GrayST & \textbf{\roundnum{83.35}} & \textbf{\roundnum{67.78}}\\ 
\hline
\end{tabular}
}
\vspace{5pt}
\caption{
Performance on CATER task 2 using 32-frame models ($T$ = 32).
We report last epoch's validation mAP using one clip. Note, GrayST uses three times as many frames as RGB or TC, which are then converted to grayscale, and TC+2 uses just two frames more.
The different sampling strategies do not impact the size of the network, \ie in the end, they all get the same number of input frames.
}
\label{tab:cater}
\end{table}

\begin{table}[t]

\centering

\begin{subtable}[t]{0.4\textwidth}
    \centering
    \caption{Something-Something-V1}
    \label{tab:somethingv1}
\resizebox{0.95\linewidth}{!}{
\begin{tabular}{|l|l|c|c|}
\hline
Model & Sampling & Top1 & Top5 \\
\hline
\hline
\multirow{4}{*}{TSN} & RGB & \roundnum{17.18} & \roundnum{42.71}\\ 
& TC & \roundnum{36.78} & \roundnum{65.31}\\ 
& TC+2 & \textbf{\roundnum{37.02}} & \textbf{\roundnum{65.63}}\\ 
& GrayST & \roundnum{35.50} & \roundnum{65.42}\\ 
\hline
\multirow{4}{*}{TRN} & RGB & \roundnum{29.69} & \roundnum{57.56}\\ 
& TC & \roundnum{35.94} & \roundnum{65.16}\\ 
& TC+2 & \textbf{\roundnum{37.43}} & \textbf{\roundnum{66.82}}\\ 
& GrayST & \roundnum{36.96} & \roundnum{66.13}\\ 
\hline
\multirow{4}{*}{MTRN} & RGB & \roundnum{30.98} & \roundnum{59.35}\\ 
& TC & \roundnum{36.63} & \roundnum{65.77}\\ 
& TC+2 & \roundnum{38.06} & \roundnum{67.52}\\ 
& GrayST & \textbf{\roundnum{38.36}} & \textbf{\roundnum{67.82}}\\ 
\hline
\multirow{4}{*}{TSM} & RGB & \roundnum{45.37} & \roundnum{74.47}\\ 
& TC & \roundnum{45.76} & \roundnum{74.74}\\ 
& TC+2 & \roundnum{46.76} & \roundnum{75.75}\\ 
& GrayST & \textbf{\roundnum{47.55}} & \textbf{\roundnum{76.65}}\\ 
\hline
\multirow{4}{*}{MVFNet} & RGB & \roundnum{46.28} & \roundnum{75.25}\\ 
& TC & \roundnum{45.72} & \roundnum{74.63}\\ 
& TC+2 & \roundnum{46.87} & \roundnum{75.78}\\ 
& GrayST & \textbf{\roundnum{47.87}} & \textbf{\roundnum{76.55}}\\ 
\hline
\multirow{4}{*}{I3D\ssymbol{1}} & RGB & \roundnum{40.52} & \roundnum{68.48}\\ 
& TC & \roundnum{39.13} & \roundnum{68.62}\\ 
& TC+2 & \roundnum{40.78} & \roundnum{69.22}\\ 
& GrayST & \textbf{\roundnum{41.10}} & \textbf{\roundnum{70.15}}\\ 
\hline
\end{tabular}}%
    \vspace{7pt}
\end{subtable}%
\qquad
\begin{subtable}[t]{0.4\textwidth}
    \centering
    \caption{Something-Something-V2}
    \label{tab:somethingv2}
\resizebox{0.95\linewidth}{!}{
\begin{tabular}{|l|l|c|c|}
\hline
Model & Sampling & Top1 & Top5\\
\hline
\hline
\multirow{4}{*}{TSN} & RGB & \roundnum{30.17} & \roundnum{60.48} \\ 
& TC & \textbf{\roundnum{51.93}} & \textbf{\roundnum{79.53}} \\ 
& TC+2 & \roundnum{51.32} & \roundnum{79.38} \\ 
& GrayST & \roundnum{50.94} & \roundnum{79.42}\\ 
\hline
\multirow{4}{*}{TRN} & RGB & \roundnum{45.68} & \roundnum{73.62} \\ 
& TC & \roundnum{51.32} & \roundnum{78.92}\\ 
& TC+2 & \textbf{\roundnum{52.30}} & \textbf{\roundnum{80.46}} \\ 
& GrayST & \roundnum{52.09} & \roundnum{79.92}\\ 
\hline
\multirow{4}{*}{MTRN} & RGB & \roundnum{46.66} & \roundnum{75.26} \\ 
& TC & \roundnum{52.02} & \roundnum{79.85} \\ 
& TC+2 & \roundnum{52.90} & \roundnum{80.94} \\ 
& GrayST & \textbf{\roundnum{52.97}} & \textbf{\roundnum{81.07}} \\ 
\hline
\multirow{4}{*}{TSM} & RGB & \roundnum{59.1} & \roundnum{85.6} \\ 
& TC & \roundnum{59.16} & \roundnum{85.45}\\ 
& TC+2 & \roundnum{59.67} & \roundnum{85.96}\\ 
& GrayST & \textbf{\roundnum{59.79}} & \textbf{\roundnum{86.16}} \\ 
\hline
\multirow{4}{*}{MVFNet} & RGB & \roundnum{59.71} & \roundnum{85.86}\\ 
& TC & \roundnum{59.62} & \roundnum{85.89}\\ 
& TC+2 & \roundnum{59.73} & \roundnum{86.05}\\ 
& GrayST & \textbf{\roundnum{60.78}} & \textbf{\roundnum{86.62}}\\ 
\hline
\end{tabular}}
    
\end{subtable}%

\caption{Validation accuracy on Something-Something with 8-frame models ($T=8$). Models marked with \nsymbol{1} use 30 clips for testing (3 spatial $\times$ 10 temporal), otherwise we report 1-clip accuracy. Our focus is on making simple 2D networks better, and thus we only report the performance of the expensive I3D on  Something-Something-V1 for comparison.
}
\label{tab:something}
\end{table}

We illustrate results for CATER Static/Camera Motion task 2 in \cref{tab:cater} and Something-Something-V1/V2 in \cref{tab:something}.
Note that the GrayST method gets to use three times more frames and TC+2 uses just two more frames than RGB or TC, while maintaining the same computational cost of the model.
We also report I3D results on Something-Something-V1, to show the impact of our sampling strategies when applied to 3D networks, see~\cref{tab:somethingv1}.
\\

\paragraphbf{TC Reordering Performance.}
Using our TC Reordering, we observe that even the simple TSN model obtains much better performance compared to using conventional RGB frames with TRN and MTRN on all datasets.
On TRN and MTRN, we observe significant improvement on most datasets, except on CATER Camera Motion task 2.
However, it has less impact when combined with TSM and MVFNet, and sometimes hurts performance. 

TC Reordering shifts the input video through its color channels in order to provide temporal information to 2D backbones. 
As a result, if a model already captures temporal information (\eg TSM and I3D), TC Reordering is less effective, but still helps in many cases.   
\\

\paragraphbf{GrayST Performance.}
We observe consistent improvement on GrayST over RGB on most datasets and all models. The first obvious reason may be that it gets to see more frames, \ie three times as many as the other strategies. 
By combining grayscale information from different points in time it enables the 2D backbones to see temporal information.
Despite utilizing more frames, this method does not increase training and inference times, with the exception of the time associated with loading the extra frames and converting them to grayscale. 
In comparison, using more RGB frames would make training and testing slower. 
\\

\paragraphbf{8-frame GrayST vs 24-frame RGB}
One might wonder if GrayST is better just because it `sees' more frames. However, it has been reported that simply increasing the number of RGB frames does not necessarily improve performance or it is very marginal~\cite{trn}, and can even decrease performance~\cite{smart} depending on the dataset. 
For example, \cite{nonlocal} showed a baseline I3D-ResNet50 network with a 32-frame input obtained 73.3\% on the Kinetics-400 dataset \cite{kinetics}. \cite{slowfast} later conducted an experiment with an 8-frame input following the same recipe for training and obtained 73.4\%.
We also conducted an ablation experiment in \cref{tab:8frame_24frame} to show that in many cases, the GrayST 8-frame (that looks at 24 frames and generates 8-frame 3-channel representation) is better than the RGB 24-frame.
A GrayST frame not only increases the number of frames without increasing training and inference cost, but also allows 2D models to see more temporal information by discarding redundant color information.
Note that TSM can make use of the information contained in the extra frames (see 8 versus 24 frames), but this benefit comes with increasing the training/testing cost by a factor of three.
\begin{table}[t]
\centering
\resizebox{0.45\linewidth}{!}{
\begin{tabular}{|l|l|c|c|}
\hline
Model & Sampling & Model Size ($T$) & Top-1\\
\hline\hline
\multirow{3}{*}{TSN} & RGB & 8 & \roundnum{17.18}\\ 
& RGB & 24 & \roundnum{18.26}\\ 
& GrayST & 8 & \textbf{\roundnum{35.50}}\\ 
\hline
\multirow{3}{*}{TRN} & RGB & 8 & \roundnum{29.69}\\ 
& RGB & 24 & \roundnum{31.68}\\ 
& GrayST & 8 & \textbf{\roundnum{36.96}}\\ 
\hline
\multirow{3}{*}{TSM} & RGB & 8 & \roundnum{45.37}\\ 
& RGB & 24 & \textbf{\roundnum{51.14}}\\ 
& GrayST & 8 & \roundnum{47.55}\\ 
\hline
\end{tabular}
}

\vspace{5pt}
\caption{Validation accuracy on Something-Something-V1 for 8 vs 24 frames. $T$ refers to the temporal size of the model, and note that the GrayST gets to access three times as many frames without increasing the model size.}
\label{tab:8frame_24frame}
\end{table}

\section{Conclusion}
We presented two novel video channel sampling strategies: TC Reordering and GrayST. The former re-orders RGB channels to increase temporal information, and the latter uses grayscale images to use more frames resulting in an increased temporal receptive field. 
In spite of the simplicity of our approaches, we observe a significant boost in performance on multiple challenging datasets. 
Importantly, these sampling strategies allow us to significantly increase the performance of existing lightweight video networks without increasing the computational cost or requiring any modifications to the underlying network architectures. 
Our hope is that this will pave the way for alternative sampling strategies. 
Future work includes developing a temporal aggregation module that is compatible with our sampling strategies for reasoning over much longer time scales. 

\pagebreak

\appendix

{\noindent\Large\textbf{\textcolor{bmv@sectioncolor}{Supplementary Material}}}

\section{Training Details}
On CATER task 2, CATER Camera Motion task 2, and Something-Something datasets, we used 2 NVIDIA RTX 3090 GPUs to train and test. 
We used 2 NVIDIA RTX 2080 Ti GPUs for the other datasets.
For CATER, we used 32 frames due to the need for long-term temporal understanding. We set the total batch size to 24 and the initial learning rate to 0.0024. 
For Something-Something, we used 8 frames, a total batch size of 64, and an initial learning rate of 0.0064.
As an exception, TRN and MTRN models are trained with one RTX 3090 GPU with half the total batch size and quarter the learning rate.

In all of the experiments, we kept the following protocols.
The videos were resized so that the shorter side becomes 224 to 336 pixels, and we performed random cropping during training.
For testing, we resized the input to have the shorter spatial side resolution of 256 and we used one center crop for CATER and Something-Something, and five crops (center and corners) and their horizontal flips for other datasets.
For I3D experiments, we followed the common practice from \cite{nonlocal} of densely sampling the video with a sampling stride of eight for RGB, three for GrayST because it samples more frames, and tested using ten evenly sampled clips throughout the video with three spatial crops of each, totaling 30 clips.
The learning rate was decayed by 0.1 when validation metrics saturated for ten epochs.
We stopped the experiments when the validation metrics saturated for 20 epochs. 
We used 16-bit precision to save memory, increase the batch size, and to train faster.

\section{Ablation Studies}
\paragraph{TC variants.} We explore some variants for ablation experiments: TC-Red, TC-RGB, and TC-ShortLong. TC-Red uses only red channels with the same frame ordering, \ie $\mathbf{x}_i^\text{TC}=(x_i^\text{R}, x_{i+1}^\text{R}, x_{i+2}^\text{R})$. This baseline allows us to measure the effect of the diversity of color information using the TC Reordering. TC-RGB uses traditional RGB-like representation with the same temporal frame ordering as the TC Reordering: $\mathbf{x}_i^\text{TC}=(x_i^\text{R}, x_{i+1}^\text{G}, x_{i+2}^\text{B})$. Intuitively, this may seem to be the best representation as this is the closest representation to the RGB representation that the model is pre-trained on. %
However, in our experiments we observe that our TC Reordering actually outperforms TC-RGB. 
Finally, TC-ShortLong replaces the last two frames that consists of a lot of duplicates, with frames having longer sampling stride instead, \ie $\mathbf{x}_7^\text{TC} = (x_3^\text{R}, x_5^\text{R}, x_7^\text{R})$ and $\mathbf{x}_8^\text{TC} = (x_4^\text{G}, x_6^\text{G}, x_8^\text{G})$ for the $T=8$ case.

\begin{table}[t]
\centering
\begin{tabular}{|l|l|c|c|}
\hline
Model & Sampling & Top-1 & Top-5 \\
\hline
\hline
\multirow{5}{*}{TSN} & RGB & \roundnum{17.18} & \roundnum{42.71}\\ 
& TC & \textbf{\roundnum{36.78}} & \textbf{\roundnum{65.31}} \\ 
& TC-RGB & \roundnum{33.61} & \roundnum{61.40} \\ 
& TC-Red & \roundnum{36.04} & \roundnum{65.18} \\ 
& TC-ShortLong & \roundnum{35.20} & \roundnum{64.44} \\ 
\hline
\multirow{5}{*}{TSM} & RGB & \roundnum{45.37} & \roundnum{74.47}\\ 
& TC & \textbf{\roundnum{45.76}} & \textbf{\roundnum{74.74}}\\ 
& TC-RGB & \roundnum{44.90} & \roundnum{74.05}\\ 
& TC-Red & \roundnum{44.71} & \roundnum{73.78} \\ 
& TC-ShortLong & \roundnum{44.75} & \roundnum{73.55} \\ 
\hline
\end{tabular}

\vspace{5pt}
\caption{
Ablation experiments of different TC Reordering strategies on Something-Something-V1. 
RGB refers to standard RGB channel sampling and TC is our proposed sampling strategy. 
}
\label{tab:TC_ablation}
\end{table}

In \cref{tab:TC_ablation} we contrast the different variants of TC Reordering. 
In the case of TSN, all variants of TC Reordering result in a significant performance increase compared to standard RGB. 
However, for TSM, only our proposed TC variant is superior. 
Perhaps counter-intuitively, TC-RGB, which maintains the RGB channel order but temporally shifts them, actually performs worse than our TC Reordering. 
TC frames consist of information from the same input color channel, which perhaps better enables it to capture local temporal changes, especially in cases where color is not informative for the task at hand.  

\paragraph{Grayscale-only} We report a grayscale-only experiment result on Something-Something-V1 in \Cref{tab:grayonly}.
We let $\mathbf{X}^\text{GO} = \left\{\mathbf{x}_1^\text{GO}, \mathbf{x}_2^\text{GO}, \dots, \mathbf{x}_T^\text{GO}\right\}$ denote a GrayOnly video clip. 
We sample $T$ grayscale frames following the sparse sampling strategy,
\begin{equation}
\mathbf{X}^g = \left\{x^g_1, x^g_2, \dots, x^g_{T}\right\}, 
\end{equation}
where $x^g_i$ is a grayscale image. 
Then, a GrayOnly frame is made by duplicating the same channel three times.
\begin{equation}
    \mathbf{x}_i^\text{GO} = (x^g_{i}, x^g_{i}, x^g_{i}).
\end{equation}
We see very little difference in performance compared to RGB.

\begin{table}[t]
\centering
\begin{tabular}{|l|l|c|c|}
\hline
Model & Sampling & Top-1 & Top-5 \\
\hline
\hline
\multirow{5}{*}{TSN} & GrayOnly & \roundnum{16.13} & \roundnum{41.16}\\
& RGB & \roundnum{17.18} & \roundnum{42.71}\\ 
& TC & \roundnum{36.78} & \roundnum{65.31}\\ 
& TC+2 & \textbf{\roundnum{37.02}} & \textbf{\roundnum{65.63}}\\ 
& GrayST & \roundnum{35.50} & \roundnum{65.42}\\
\hline
\end{tabular}

\vspace{5pt}
\caption{
Grayscale-only result on Something-Something-V1. Surprisingly, the performance is very close to that of RGB, and our methods utilize this fact to make the models further capture temporal information.
}
\label{tab:grayonly}
\end{table}

\section{Only Time Can Tell dataset}

We show results on OnlyTimeCanTell-Temporal dataset~\cite{onlytime} in \Cref{tab:octc}. The dataset consists of 50 classes which require extensive temporal reasoning, taken from the Kinetics-400 and Something-Something-V1 dataset. Our methods outperformed the baselines by a significant margin on this dataset.

\begin{table}[t]
\centering
\begin{tabular}{|l|l|c|c|}
\hline
Model & Sampling & Top-1 & Top-5 \\
\hline
\hline
\multirow{3}{*}{TSN} & RGB & \roundnum{50.69} & \roundnum{84.62}\\ 
& TC+2 & \roundnum{65.85} & \roundnum{91.99}\\ 
& GrayST & \textbf{\roundnum{66.43}} & \textbf{\roundnum{92.67}}\\
\hline
\multirow{3}{*}{TSM} & RGB & \roundnum{71.70} & \roundnum{94.22}\\ 
& TC+2 & \roundnum{73.32} & \roundnum{94.04}\\ 
& GrayST & \textbf{\roundnum{73.68}} & \textbf{\roundnum{95.02}}\\
\hline
\end{tabular}

\vspace{5pt}
\caption{
Evaluation of our methods on the OnlyTimeCanTell-Temporal dataset.
}
\label{tab:octc}
\end{table}

\section{Limitations}
Despite the positive results, we also find some limitations of the proposed approaches, which we hope can lead to interesting future work.
\\

\paragraph{Datasets Requiring Less Temporal Reasoning.}
Numerous datasets have significant object and scene bias making even TSN perform very similar to powerful 3D networks. In such cases, we found that the sampling strategies do not result in improved performance. \Cref{tab:diving,tab:ucf} show such cases on Diving-48 \cite{diving48} and UCF-101 \cite{ucf101} datasets. 

Interestingly, Diving-48 V2 shows a decrease in performance with the GrayST input. The implication of this is that color information is important on this dataset, and with grayscale images it is difficult to distinguish divers of interest from the background. Despite the fact that the dataset is said to be ``temporally-heavy'', our experiment showed that the gap between 2D network and the state-of-the-art 3D network with double the number of frames and backbone depth is marginal.

The UCF-101 labels are biased towards object and scene information. As a result the performance of TSM is almost identical to that of TSN. The result did not show strong pattern but in general RGB seems to be preferable in this case.

\begin{table}[t]
\centering
\begin{tabular}{|l|l|c|l|c|}
\hline
Model & Backbone & \#frame & Sampling & Top-1\\
\hline\hline
\multirow{3}{*}{TSN} & \multirow{3}{*}{ResNet50} & \multirow{3}{*}{8} & RGB & \textbf{\roundnum{75.43}} \\ 
&&& TC & \roundnum{74.97} \\ 
&&& GrayST & \roundnum{73.50} \\ 
 \hline
SlowFast\ssymbol{1} & ResNet101 & 16 & RGB & 77.6 \\ 
\hline
\end{tabular}

\vspace{5pt}
\caption{Evaluation on Diving-48-V2. The result marked with \nsymbol{1} is from~\cite{bertasius2021space}, which uses 30 clips for testing (3 spatial $\times$ 10 temporal).}
\label{tab:diving}
\end{table}

\begin{table}[t]
\centering
\begin{tabular}{|l|l|l|c|c|}
\hline
Model & Backbone & Sampling & Top-1 & Top-5\\
\hline\hline
\multirow{3}{*}{TSN} & \multirow{3}{*}{ResNet50} & RGB & \roundnum{83.58} & \textbf{\roundnum{96.11}} \\ 
&& TC & \roundnum{83.61} & \roundnum{96.01} \\ 
&& GrayST & \textbf{\roundnum{84.69}} & \roundnum{95.88}\\ 
\hline
\multirow{3}{*}{TRN} & \multirow{3}{*}{ResNet50} & RGB & \textbf{\roundnum{84.56}} & \textbf{\roundnum{96.62}} \\ 
&& TC & \roundnum{81.60} & \roundnum{95.29} \\ 
&& GrayST & \roundnum{82.16} & \roundnum{95.51} \\ 
\hline
\multirow{3}{*}{TSM} & \multirow{3}{*}{ResNet50} & RGB & \textbf{\roundnum{83.74}} & \roundnum{95.98} \\ 
&& TC & \roundnum{81.42} & \roundnum{95.48} \\ 
&& GrayST & \roundnum{83.58} & \textbf{\roundnum{96.11}} \\ 
 \hline
\end{tabular}

\vspace{5pt}
\caption{8-frame evaluation on UCF-101 split 1.}
\label{tab:ucf}
\end{table}

\paragraph{Camera Motion.}
We found that TC Reordering combined with TRN and TSM negatively impacts performance on the CATER Camera Motion dataset. Note that the models make use of the temporal ordering of frames while TSN does not.
Again, we still saw improvement with the GrayST method on this dataset. 
Judging from the fact that TC Reordering only hurts the temporal models, we think that this ordering plays a critical role in heavy camera motion scenarios.

Additionally, this dataset requires 3D-geometric understanding as the camera orbits around the objects substantially, making stationary objects look like they are sliding. The strength of TC Reordering is in cases where the model can analyze the motion information, but the large camera motion likely confuses the model when trying to understand what is the action of interest and what is the camera motion.

\end{document}